\documentclass{article}
\usepackage[main, final]{neurips_2025}

\usepackage{etoc}
\usepackage[utf8]{inputenc}
\usepackage[T1]{fontenc}
\usepackage{url}
\usepackage{booktabs}
\usepackage{amsfonts}
\usepackage{nicefrac}
\usepackage{microtype}

\usepackage{enumitem}
\usepackage{graphicx}
\usepackage{amsmath}
\usepackage[normalem]{ulem}
\usepackage{multirow}
\usepackage{caption}
\usepackage{wrapfig}
\usepackage{stmaryrd}
\usepackage[table]{xcolor}

\usepackage[colorlinks, linkcolor=mydarkred, citecolor=myblue]{hyperref}
\definecolor{mydarkblue}{rgb}{0,0.08,0.45}
\definecolor{mydarkred}{rgb}{0.6,0,0}
\definecolor{myblue}{HTML}{268BD2}
\definecolor{mygreen}{HTML}{658354}
\definecolor{results}{RGB}{220, 230, 240}

\usepackage[most]{tcolorbox}

\title{GeoRanker: Distance‑Aware Ranking for Worldwide Image Geolocalization}

\author{
    Pengyue Jia$^{1,2}$, Seongheon Park$^2$, Song Gao$^3$, \textbf{Xiangyu Zhao}$^1$, \textbf{Sharon Li}$^2$\\ 
    $^1$Department of Data Science, City University of Hong Kong, \\$^2$Department of Computer Sciences, University of Wisconsin-Madison  \\
    $^3$Department of Geography, University of Wisconsin-Madison \\
    \texttt{jia.pengyue@my.cityu.edu.hk,sharonli@cs.wisc.edu}
}

\begin{document}
\maketitle
\begin{abstract}
Worldwide image geolocalization—the task of predicting GPS coordinates from images taken anywhere on Earth—poses a fundamental challenge due to the vast diversity in visual content across regions. While recent approaches adopt a two-stage pipeline of retrieving candidates and selecting the best match, they typically rely on simplistic similarity heuristics and point-wise supervision, failing to model spatial relationships among candidates. In this paper, we propose \textbf{GeoRanker}, a distance-aware ranking framework that leverages large vision-language models to jointly encode query–candidate interactions and predict geographic proximity. In addition, we introduce a \emph{multi-order distance loss} that ranks both absolute and relative distances, enabling the model to reason over structured spatial relationships. To support this, we curate {GeoRanking}, the first dataset explicitly designed for geographic ranking tasks with multimodal candidate information. GeoRanker achieves state-of-the-art results on two well-established benchmarks (IM2GPS3K and YFCC4K), significantly outperforming current best methods. We also release our code, checkpoint, and dataset online\footnote{\url{https://github.com/Applied-Machine-Learning-Lab/GeoRanker}} for ease of reproduction.
\end{abstract}
\section{Introduction}
\setlength{\intextsep}{-7pt}
\setlength{\columnsep}{10pt}
\begin{wrapfigure}{r}{0.30\textwidth}
\vspace{0mm}
\centering
    \includegraphics[width=\linewidth]{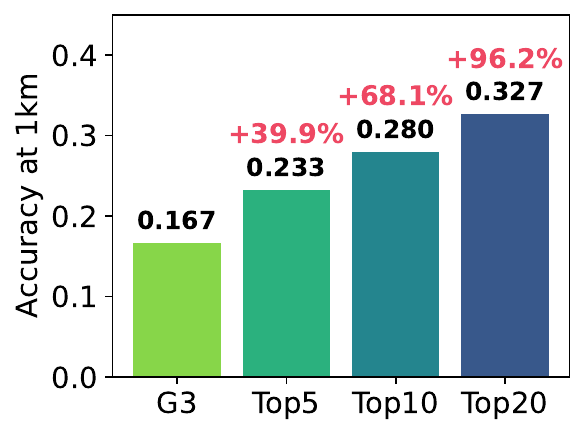}
    \caption{Accuracy at 1km error threshold for G3 (Current SOTA) vs. the best candidate within top‑k retrieved results.
    }
    \label{fig:motivation}
\end{wrapfigure}
Worldwide geolocalization~\cite{vo2017revisiting,wilson2021visual} refers to the task of predicting the GPS coordinates of images captured anywhere on Earth. Unlike approaches constrained to specific cities or regions~\cite{noh2017large,cao2020unifying,tan2021instance,lee2022correlation,shao2023global}, global geolocalization poses significantly greater challenges due to the immense diversity in visual content, ranging from natural landscapes and climatic variations to architectural differences and cultural markers~\cite{vivanco2023geoclip,haas2024pigeon,zhou2024img2loc}. Despite these complexities, accurate global geolocalization holds broad practical relevance, with a wide range of applications including criminal investigations~\cite{liu2024image}, navigation systems~\cite{jay2025evaluating}, and environmental monitoring~\cite{wang2024llmgeo}.

Recent state-of-the-art approaches~\cite{vivanco2023geoclip,haas2024pigeon,zhou2024img2loc,jia2024g3} typically follow a two-stage pipeline: (1) retrieving and generating a set of candidates based on a global database, and (2) selecting the top match as the predicted geolocation. As shown in Figure~\ref{fig:motivation}, although the current SOTA model (G3~\cite{jia2024g3}, which ranks candidates via GPS location-image embedding similarity) achieves 16.7\% accuracy at the 1km error threshold on IM2GPS3K, better candidates often exist among the top‑k retrieved results.
This suggests that one can retrieve reasonably high-quality candidates, yet the final prediction accuracy hinges on the second stage---the model’s ability to compare spatial relevance and select the most plausible candidate. Currently, the candidate selection is often limited by naïve heuristics such as cosine similarity~\cite{vivanco2023geoclip,jia2024g3}, which generally encode the query image and candidates \emph{independently}, without modeling their mutual spatial relationships or allowing rich interactions. As a result, these methods frequently struggle to distinguish between visually similar yet geographically distant scenes. 
Furthermore, existing training objectives primarily focus on point-wise similarity between individual images and locations~\cite{vivanco2023geoclip,haas2024pigeon,seo2018cplanet}, overlooking the rich spatial relationships among candidates—such as the spatial dependence (i.e., Tobler's First Law of Geography~\cite{miller2004tobler}) and relative distances between them—which are crucial for geolocalization.

To address these limitations, we propose \textbf{GeoRanker}, a \emph{distance-aware ranking framework} designed to model spatial relationships among candidate locations. Rather than relying on independent similarity scores, GeoRanker models the interaction between the query image and each candidate through a large vision-language model (LVLM), which captures rich spatial semantics via cross-modal alignment,
and learns a scalar distance score that reflects their geographic proximity. Central to our approach is a multi-order distance optimization objective that ranks not only the absolute distances between the query and individual candidates (first-order supervision), but also the relative differences between candidate distances (second-order supervision). This formulation allows the model to learn both which candidate is closest and how much closer it is compared to others, capturing rich spatial structure that naïve heuristics overlook.
Through this design, GeoRanker transforms the geolocalization task from one of isolated similarity matching to one of structured spatial reasoning.

To support this training paradigm, we construct GeoRanking, a new dataset that provides spatially diverse candidate sets for each query. Each candidate is annotated with GPS coordinates, textual descriptions (e.g., city, country), and image data. To the best of our knowledge, this is the first ranking dataset specifically designed for modeling spatial relationships among geographic entities. We believe this effort will significantly contribute to advancing research in related domains. We validate the effectiveness of GeoRanker through extensive experiments on two widely used benchmarks: IM2GPS3K~\cite{hays2008im2gps} and YFCC4K~\cite{thomee2016yfcc100m}. GeoRanker achieves \emph{state-of-the-art} performance across all geographic thresholds. For example, on IM2GPS3K, it improves street-level (1km) accuracy by +\textbf{12.9}\% over the current best method~\cite{jia2024g3}, and on YFCC4K, it yields an +\textbf{37.3}\% improvement at the same threshold. Our model also consistently outperforms existing approaches at coarser scales (25km, 200km, 750km, 2500km), highlighting its robustness across granularities. Ablation studies confirm that both components of our multi-order distance loss—first-order and second-order supervision—contribute to improved accuracy. We also conduct comprehensive ablations to understand the impact of various hyperparameter choices, leading to an improved understanding of our framework. Our key contributions are summarized as follows:

\begin{enumerate}[leftmargin=*]
\item We introduce GeoRanker, a distance-aware ranking framework that models spatial relationships among candidate locations using a multi-order distance loss and large vision-language models.

\item We construct GeoRanking, the first dataset tailored for spatial ranking tasks, with rich multimodal annotations spanning GPS coordinates, textual descriptions, and image data---facilitating future research in related fields.

\item We achieve state-of-the-art performance on two well-established public geolocalization benchmarks, with substantial gains at fine-grained localization levels, and demonstrate the effectiveness of our approach through comprehensive ablations.
\end{enumerate}

\section{Related Work}

\textbf{Image Geolocalization.} Worldwide geolocalization~\cite{sarkar2024gomaa,astruc2024openstreetview,dufour2025around,dou2024gaga} lies at the intersection of geography and computer vision, and is a core topic in GeoAI~\cite{mai2024opportunities,janowicz2020geoai,li2024georeasoner} and spatial data mining~\cite{wang2020deep,liang2025foundation}. Existing methods for worldwide geolocalization can be grouped into three main categories: classification-based, retrieval-based, and RAG-based approaches. (1) \emph{Classification-based} methods~\cite{weyand2016planet,seo2018cplanet,pramanick2022world,clark2023we,haas2024pigeon} approach the task by partitioning the Earth’s surface into discrete Geo-grids and predicting the index of the grid that contains the image location. The final output is typically the center coordinate of the predicted grid. While these methods offer scalability, they may incur large errors when the true location lies far from the grid center, even if the grid prediction is correct. (2) \emph{Retrieval-based} methods~\cite{zhu2022transgeo,lin2022joint,zhang2023cross,workman2015wide,liu2019lending,zhu2021vigor} cast geolocalization as a similarity search problem. These methods either use a database of geotagged images~\cite{yang2021cross,zhu2022transgeo,tian2017cross,shi2020looking,zhu2023r2former} or a gallery of GPS points~\cite{vivanco2023geoclip}, returning the coordinates of the most similar entries to the input image as the prediction. However, these methods fail to capture the complex spatial relationships between the query image and candidate locations, making it difficult to reliably identify the most accurate match from the candidate pool. (3) \emph{RAG-based} methods~\cite{zhou2024img2loc,jia2024g3} first retrieve a set of candidate locations similar to the query image from the database, then construct a prompt that integrates both the query and candidate information. This prompt is passed to an LVLM to generate a plausible GPS location. 
In contrast to the above approaches, our proposed \textit{Distance-Aware Ranking} framework, GeoRanker, focuses specifically on the candidate ranking stage. By explicitly modeling the complex spatial relationships between the query image and candidate geographic entities, an aspect overlooked by prior work, our approach offers more reliable candidate selection and leads to improved geolocalization performance at the global scale. 

\textbf{Learning to Rank.} Learning-to-rank~\cite{cao2007learning} (LTR) is a fundamental research direction in information retrieval~\cite{liu2009learning} and recommender systems~\cite{karatzoglou2013learning}, primarily used to train ranking models that refine the order of retrieved candidates based on a given query~\cite{kabir2024survey}. Depending on their modeling and optimization strategies, LTR methods are typically categorized into three types: pointwise, pairwise, and listwise approaches. Pointwise methods~\cite{bell2018title,dadaneh2020arsm} take the ranking task as a regression or classification problem by assigning a relevance score or label to each query–candidate pair independently. This approach is simple and straightforward, yet it overlooks the relative relationships among candidates. Pairwise methods~\cite{tagami2013ctr,cerrato2020fair,jia2021pairrank} model the relative preferences between pairs of candidates for the same query. Pairwise methods encourage the model to assign higher scores to positive candidates while penalizing negative ones, learning the relative preferences between different candidates. Listwise methods~\cite{burges2010ranknet,xia2008listwise,cao2007learning,xu2007adarank} can be seen as an extension of pairwise approaches, as they consider the entire list of candidates associated with a query and optimize a loss function that directly reflects the overall quality of the ranking.
Our approach builds on the LTR foundation but adapts it to spatial ranking by explicitly modeling and optimizing distance-aware relationships between candidates.

\section{Methodology}

\begin{figure}
    \centering
    \includegraphics[width=\linewidth]{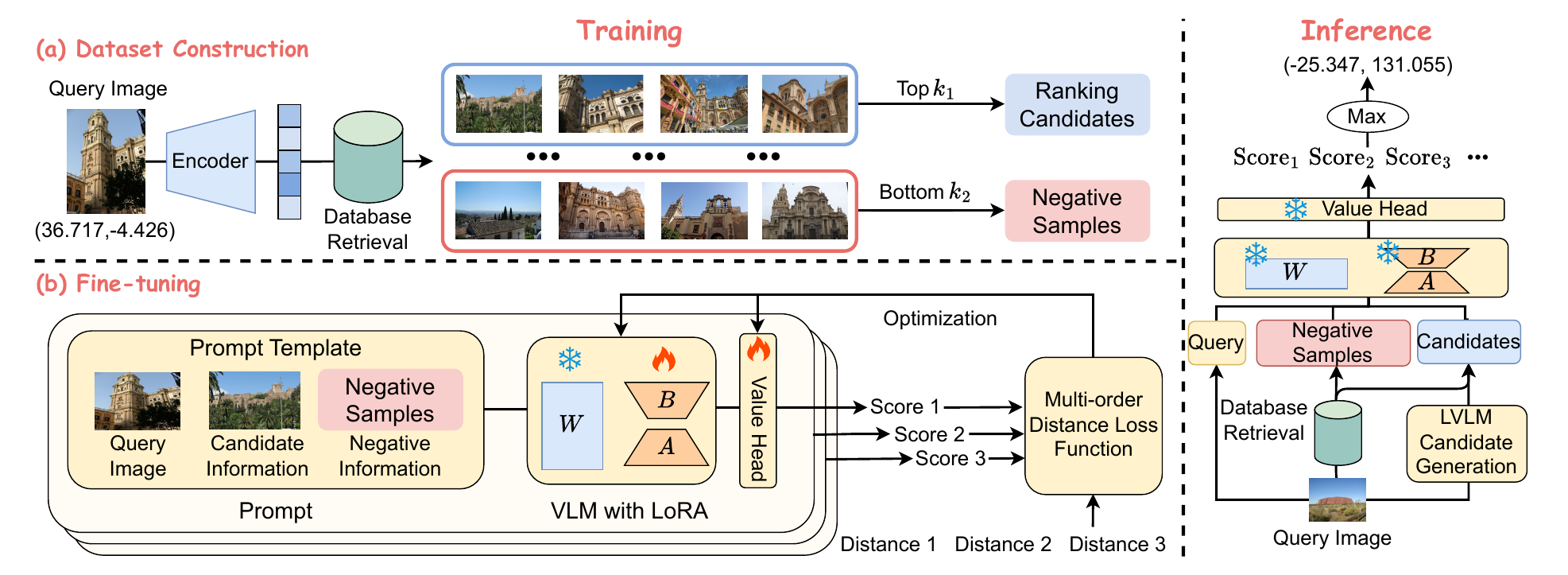}
    \caption{Overview of the Distance-aware Ranking framework--GeoRanker.
    }
    \label{fig:overview}
\end{figure}

In this section, we introduce GeoRanker, a \textit{Distance-aware Ranking} method for geolocalization. An overview of the framework is illustrated in Figure~\ref{fig:overview}, which consists of two main phases: training and inference. The training phase begins with dataset construction, detailed in Section~\ref{sec:dataset_construction}, where we describe how the GeoRanking dataset is built to support the training of GeoRanker. We then present the model architecture and optimization strategy of GeoRanker in Section~\ref{sec:georanker} and Section~\ref{sec:optimization}. Finally, during inference, GeoRanker selects the most appropriate candidate as the prediction by scoring the spatial relationships between the query image and a set of candidates (Section~\ref{sec:inference}).

\subsection{GeoRanking Dataset Construction} \label{sec:dataset_construction}

\textbf{Database.} Following prior work~\cite{jia2024g3}, we adopt the MP16-Pro multimodal dataset~\cite{jia2024g3} as our database and encode each sample into vectors. Each candidate entry includes GPS, textual descriptions (city, country, etc.), and image data. In a candidate database $\mathcal{C}=\{c_1, c_2,c_3,\cdots,c_{M}\}$, each candidate $c_m$ is encoded into a feature vector $\mathbf{v}_{c_m}=\text{concat}(\text{Encoder}_c^{\text{gps}}(c_m^{\text{gps}}),\text{Encoder}_c^{\text{text}}(c_m^{\text{text}}),\text{Encoder}^{\text{img}}{(c_m^{\text{img}})})$, where $c_m^{\text{gps}},c_m^{\text{text}},c_m^{\text{img}}$ represent the GPS, textual, and image modalities of the $m$-th candidate and $\text{Encoder}_c^{\text{gps}},\text{Encoder}_c^{\text{text}},\text{Encoder}^{\text{img}}$ are their corresponding modality-specific encoders. 

\textbf{Retrieval.} Since the input query image $q$ is a single modality (image), we design its representation to be compatible with the multimodal candidate vectors for similarity matching. The query image is encoded as: $\mathbf{v}_q=\text{concat}(f_{\text{img}\shortrightarrow \text{gps}}(\text{Encoder}^{\text{img}}(q)),f_{\text{img}\shortrightarrow \text{text}}(\text{Encoder}^{\text{img}}(q)),\text{Encoder}^{\text{img}}(q))$, 
where $f_{\text{img}\shortrightarrow \text{gps}}(\cdot)$ and $f_{\text{img}\shortrightarrow \text{text}}(\cdot)$ are adapter layers that project the visual features into the GPS and textual embedding spaces, respectively.
These adapters, along with the GPS encoder, are trained using an InfoNCE loss~\cite{oord2018representation} to align the query and candidate representations, as in G3~\cite{jia2024g3}. The remaining encoders are initialized with pretrained weights and kept frozen during training. To retrieve candidates for the query image, we compute the cosine similarity between $ \mathbf{v}_q $ and each candidate's representation $ \mathbf{v}_{c_m} $, and select the top-$ N $ candidates with the highest similarity scores to form the candidate set:
$$\mathcal{C}^{\prime} = \{c_{1}^{\prime}, c_{2}^{\prime}, \dots, c_{N}^{\prime} \mid \text{sim}(q, c_{1}^{\prime}) \geq \text{sim}(q, c_{2}^{\prime}) \geq \dots \geq \text{sim}(q, c_{N}^{\prime})\},$$
where the similarity function is defined as $\text{sim}(q, c_m) = (\mathbf{v}_q \cdot \mathbf{v}_{c_m})/(\|\mathbf{v}_q\| \cdot \|\mathbf{v}_{c_m}\|)$. The query image, along with the retrieved candidates will be used for training the GeoRanker, which we described next.

\subsection{GeoRanker} \label{sec:georanker}

Figure~\ref{fig:overview} (b) shows the overview of GeoRanker. Existing methods~\cite{vivanco2023geoclip,jia2024g3} typically model the query image and candidate geographic entities separately, embedding them into a shared representation space via independent encoders. The final prediction is based on similarity scores between these representations. However, such designs fail to capture the rich spatial interactions between the query and candidates, resulting in a decoupled modeling process that limits the accuracy of worldwide geolocalization. To address this issue, we propose GeoRanker, a distance-aware ranking model designed to capture the spatial relationships between query–candidate pairs. 
Specifically, the query and candidates are assembled into a prompt following a predefined template. These inputs are then processed by an LVLM to model the complex interactions between the query and candidate. Finally, a linear value head maps the hidden states to a scalar score that reflects the geographic distance between the query image and the candidate location.

\textbf{GeoRanking dataset and prompt construction.} To support the ranking model training, we select the top-$ k_1 $ candidates from the candidate set $\mathcal{C}'$ as ranking candidates, denoted by $\mathcal{C}_{\text{rc}}=\{c_1^{\prime},c_2^{\prime},\dots,c_{k_1}^{\prime}\}$. We then take the last $k_2$ candidates in $\mathcal{C}^{\prime}$ to form the negative set $\mathcal{C}_{\text{neg}}$, which provides additional contextual diversity and helps the model understand the relative relevance of candidate locations. Thus, each sample in the GeoRanking dataset is represented as a triplet: \(\{q, \mathcal{C}_{\text{rc}}, \mathcal{C}_{\text{neg}}\}\), where \( q \) is the query image, \(\mathcal{C}_{\text{rc}}\) contains the candidates to be ranked, and \(\mathcal{C}_{\text{neg}}\) provides hard negatives to enhance ranking discrimination.
Each triplet is formatted using a structured prompt template:
\begin{tcolorbox}[colback=gray!5!white,colframe=gray!75!black]
\textcolor{myblue}{\{query image\}} How far is this place from latitude:  \textcolor{myblue}{\{candidate latitude\}}, longitude: \textcolor{myblue}{\{candidate longitude\}}, \textcolor{myblue}{\{candidate textual descriptions\}}, \textcolor{myblue}{\{candidate image\}}? Negative examples: \textcolor{myblue}{\{negative information\}}.
\end{tcolorbox}
The construction process can be formalized as: $\mathbf{x} = \text{Prompt}(q, \mathcal{C}_{\text{rc}}, \mathcal{C}_{\text{neg}},p)$, where $p$ denotes the prompt template. Note that, to reduce GPU memory consumption, we represent negative samples using only their textual GPS coordinates and textual descriptions.

\textbf{Model architecture.} The constructed input $\mathbf{x}$ is fed into LVLM to encode both visual and textual modalities and to capture the spatial interactions between the query and candidate.
To enhance the model's representation capacity while maintaining training efficiency, we insert LoRA (Low-Rank Adaptation)~\cite{hu2022lora} modules into the intermediate layers of the LVLM backbone during training. 

We use the hidden states corresponding to the final position token as the joint representation of the input. A lightweight value head, implemented as a single linear layer without bias, is then applied to map this representation to a scalar score. This score serves as an estimate of the geographic distance between the query image and the candidate location (higher score corresponds to a smaller distance):
\begin{equation}
    s = \mathbf{w}^\top \mathbf{h}_{\text{final}}, \quad \text{where } \mathbf{h}_{\text{final}} = \text{LVLM}(\mathbf{x})_{[-1]}
\end{equation}
where $s \in \mathbb{R}$ denotes the final score, $\mathbf{w} \in \mathbb{R}^{\text{dim}}$ and $\mathbf{h}_{\text{final}} \in \mathbb{R}^{\text{dim}}$ are the weight matrix of the value head and the final position token's representation, $\text{dim}$ is the dimension of the last hidden states. In addition, $\text{LVLM}(\cdot)$ denotes the large vision-language model used to encode the input $\mathbf{x}$.

\subsection{GeoRanker: Optimization with Multi-Order Distance Objective} \label{sec:optimization}

Existing geolocalization training methods typically focus on point-wise image-to-location similarity, without modeling the spatial relationships among candidate locations. To address this limitation, we propose a multi-order distance optimization objective to train GeoRanker. Our objective incorporates both the first-order distances between the query and each candidate, and the second-order relationships, defined as the relative differences between first-order distances, to guide the model during training.

\textbf{First-order distance loss.} We optimize the first-order distance ranking using a partial Plackett-Luce (PL) loss~\cite{plackett1975analysis,luce1959individual}. Given $k_1$ candidates with predicted scores $\{s_1, s_2, \dots, s_{k_1}\}$, we first sort the corresponding geodesic distances $\{d_1, d_2, \dots, d_{k_1}\}$ in ascending order to obtain an index permutation $\pi$ such that $d_{\pi(1)} < d_{\pi(2)} < \dots < d_{\pi(k_1)}$. We then use the reordered scores $\{s_{\pi(1)}, s_{\pi(2)}, \dots, s_{\pi(k_1)}\}$ to compute the loss. Let $K^{(1)} \leq k_1$ be a hyperparameter controlling how many top-ranked candidates are included in the objective. For each sample, the partial Plackett-Luce loss is defined as:
\begin{equation}
\mathcal{L}_{\text{PL}}^{(1)} = - \frac{1}{K^{(1)}}\sum_{i=1}^{K^{(1)}} \log \frac{\exp(s_{\pi(i)})}{\sum_{j=i}^{k_1} \exp(s_{\pi(j)})}
\end{equation}
\textbf{Second-order distance loss.}  
To capture the relative spatial differences among candidates, we introduce a second-order distance loss based on pairwise distance gaps. This objective supervises the \emph{ranking of first-order distance differences}, encouraging the model to assign higher score differences to candidate pairs that are more distant in geolocation. Specifically, we first compute all pairwise first-order differences in distances and predicted scores:
\begin{equation}
    \Delta d_{i,j} = d_{\pi(i)} - d_{\pi(j)}, \quad \Delta s_{i,j} = s_{\pi(i)} - s_{\pi(j)}, \quad \text{for } 1 \leq i < j \leq k_1
\end{equation}
This results in $P=\frac{k_1(k_1 - 1)}{2}$ pairs. We sort the distance differences $\Delta d_{i,j}$ in ascending order (so larger spatial gaps appear earlier), and apply the same permutation to the score differences $\Delta s_{i,j}$, resulting in an ordered sequence $\Delta s_{(1)}, \dots, \Delta s_{(P)}$.

Let $K^{(2)}$ be a hyperparameter that specifies the number of top-ranked pairs included in the loss. We define $K^{(2)} = \frac{[(k_1 - 1) + (k_1 - K^{(1)})] \times K^{(1)}}{2}$, which ensures that the second-order loss focuses on candidate pairs where at least one candidate is involved in the first-order loss computation.
The second-order partial PL loss is then computed as:
\begin{equation}
    \mathcal{L}_{\text{PL}}^{(2)} = - \frac{1}{K^{(2)}} \sum_{i=1}^{K^{(2)}} \log \frac{\exp(\Delta s_{(i)})}{\sum_{j=i}^{P} \exp(\Delta s_{(j)})}
\end{equation}
This formulation encourages the model to preserve the ordering of spatial gaps in the score space, so that larger geographic differences lead to larger score gaps.

\textbf{Joint optimization.}
We jointly optimize the model with both the first-order and second-order objectives. The total loss is defined as a weighted sum of the two components:
\begin{equation}
\mathcal{L}_{\text{total}} = \lambda \cdot \mathcal{L}_{\text{PL}}^{(1)} + (1-\lambda) \cdot \mathcal{L}_{\text{PL}}^{(2)} 
\end{equation}
where $\lambda$ is the weighting coefficient that balances the contribution of the first-order and second-order distance losses, respectively. We will ablate the impact of key hyperparameters in Section~\ref{sec:ablation}.

\subsection{Inference} \label{sec:inference} 
During inference, GeoRanker integrates both retrieved candidates from a database and generated candidates from an LVLM, following prior work~\cite{zhou2024img2loc,jia2024g3}. Given a query image $q$, we first retrieve a set of candidates $\mathcal{C}_{\text{r}}$ and collect contextual negative samples $\mathcal{C}_{\text{neg}}$. Simultaneously, the query $q$ is passed through an LVLM to generate a new set of candidates $\mathcal{C}_{\text{g}}$, referred to as generated candidates. The prompt for generating candidates is detailed in \textbf{Appendix~\ref{sec:appendix_generation_prompt}}. We then form query–candidate pairs by combining $q$ with each candidate $c \in \mathcal{C}_{\text{r}} \cup \mathcal{C}_{\text{g}}$, and feed these inputs into GeoRanker to obtain a set of distance scores: $s_c = \text{GeoRanker}(q, c, \mathcal{C}_{\text{neg}}), \quad \forall c \in \mathcal{C}_{\text{r}} \cup \mathcal{C}_{\text{g}}$.
Finally, we select the candidate with the highest score and use its GPS coordinates as the prediction:
\begin{equation}
    \hat{c} = \arg\max_{c \in \mathcal{C}_{\text{r}} \cup \mathcal{C}_{\text{g}}} s_c
\end{equation}
It is worth noting that the generated candidates typically lack additional modalities such as textual descriptions and images. As a result, we use only their GPS coordinates during inference.

\section{Expeirments}

\subsection{Setup} \label{sec:setup}
\textbf{Dataset and evaluation metrics.}
We use MP16-Pro from prior work~\cite{jia2024g3} as our database in constructing the GeoRanking dataset. For evaluation, we follow previous work~\cite{vivanco2023geoclip,zhou2024img2loc,jia2024g3} and assess performance on two widely used public benchmarks IM2GPS3K~\cite{hays2008im2gps} and YFCC4K~\cite{thomee2016yfcc100m}. The evaluation metric reports the percentage of predictions whose geodesic distance to the ground-truth coordinates falls within a set of thresholds: 1km, 25km, 200km, 750km, and 2500km.

\textbf{GeoRanking dataset.}
In this work, we construct the first ranking dataset for modeling distances between geographic entities. Specifically, for each query image, we retrieve a set of candidates based on embedding similarity. Each candidate is associated with GPS coordinates, textual descriptions (e.g., city, country), and image data. These candidates serve as input for subsequent ranking models. In total, we construct \textbf{100k samples}, resulting in \textbf{2 million query–candidate pairs}. By releasing this dataset, we aim to support progress in geolocalization and related research areas such as GeoAI, information retrieval, and LVLM. Example entries are provided in the \textbf{Appendix~\ref{sec:appendix_georanking}} for reference.

\textbf{Implementation details.} During training, we retrieve 20 candidates from the database for each query. The top-7 are used as retrieval candidates, while the bottom-5 serve as negative samples. The vision encoder and text encoder are pretrained models from CLIP~\cite{radford2021learning}. The GPS encoder is initialized with weights from GeoCLIP~\cite{vivanco2023geoclip} and then fine-tuned.
We use Qwen2-VL-7b-Instruct\footnote{\url{https://huggingface.co/Qwen/Qwen2-VL-7B-Instruct}} as the LVLM backbone in GeoRanker. For LoRA fine-tuning, we target the \texttt{q\_proj}, \texttt{k\_proj}, and \texttt{v\_proj} modules, with a rank of 16, scaling factor of 32, and LoRA dropout of 0.05. GeoRanker is fine-tuned with AdamW~\cite{loshchilov2017decoupled} optimizer with a learning rate of 1e-4, a batch size of 4, and for 1 epoch. For joint optimization, we set the weighting coefficient $\lambda=0.7$, and $K^{(1)}=1$. All experiments are conducted using Pytorch on 4 NVIDIA L40S GPUs. During inference, following~\cite{zhou2024img2loc, jia2024g3}, we use GPT4V\footnote{\url{https://openai.com/}} as the LVLM for candidate generation. We use $\rvert C_r \rvert=12$ retrieved candidates and $\rvert C_g \rvert=3$ generated candidates for IM2GPS3K and $\rvert C_r \rvert=14, \rvert C_g \rvert=5$ for YFCC4K. Additional details regarding the training environment and runtime are provided in \textbf{Appendix~\ref{sec:appendix_hyper}}.

\textbf{Baselines.} To evaluate the effectiveness of our approach, we conduct comprehensive experiments and compare it against 11 baselines: [L]kNN, sigma=4~\cite{vo2017revisiting}, PlaNet~\cite{seo2018cplanet}, CPlaNet~\cite{seo2018cplanet}, ISNs~\cite{muller2018geolocation}, Translocator~\cite{pramanick2022world}, GeoDecoder~\cite{clark2023we}, GeoCLIP~\cite{vivanco2023geoclip}, Img2Loc~\cite{zhou2024img2loc}, PIGEON~\cite{haas2024pigeon}, G3~\cite{jia2024g3}, including the state-of-the-art. A detailed description of each baseline is provided in \textbf{Appendix~\ref{sec:appendix_baseline}}.

\subsection{Main Results}

As shown in Table~\ref{tab:overall}, GeoRanker achieves state-of-the-art performance across all evaluation thresholds. For example, on IM2GPS3K, it improves the most challenging street-level accuracy by \textbf{12.9}\% over the best baseline G3, and on YFCC4K, it achieves an \textbf{37.3}\% relative gain at the same threshold.
Among the baselines, GeoCLIP~\cite{vivanco2023geoclip}, Img2Loc~\cite{zhou2024img2loc}, PIGEON~\cite{haas2024pigeon}, and G3~\cite{jia2024g3} exhibit relatively strong performance due to classification-based methods are limited by systemic biases from fixed candidate grids.
Compared to these stronger baselines, our method GeoRanker achieves superior results by explicitly modeling the spatial relationship between each query–candidate pair using a multi-order distance optimization objective. This enables the model to accurately identify the geographically closest candidate as the prediction, further enhancing geolocalization accuracy.
In summary, our approach achieves state-of-the-art performance across all datasets and metrics, demonstrating its effectiveness and superiority. Furthermore, \textbf{Appendix~\ref{sec:appendix_error_thresholds}} presents representative examples across various error thresholds to offer intuitive insights into the distribution of query images at different localization accuracies. \textbf{Appendix~\ref{sec:ranking_metrics}} validates the ranking capability of GeoRanker using standard learning-to-rank metrics.

\begin{table}
\centering
\caption{\textbf{Main results} on IM2GPS3K and YFCC4K. For all metrics, higher is better. The best-performing results are highlighted in \textbf{bold}, while the second-best results are \underline{underlined}. 
$\Delta$ represents the relative improvement of our method over the best baseline.}
\label{tab:overall}
\resizebox{\linewidth}{!}{
\begin{tabular}{cccccccccccc} 
\toprule
\multicolumn{2}{c}{\multirow{2}{*}{\textbf{Methods}}} & \multicolumn{5}{c}{IM2GPS3K}                                                                                                                                                                                                                                                          & \multicolumn{5}{c}{YFCC4K}                                                                                                                                                                                                                                                             \\ 
\cmidrule[\heavyrulewidth]{3-12}
\multicolumn{2}{c}{}                         & \begin{tabular}[c]{@{}c@{}}Street\\1km\end{tabular} & \begin{tabular}[c]{@{}c@{}}City\\25km\end{tabular} & \begin{tabular}[c]{@{}c@{}}Region\\200km\end{tabular} & \begin{tabular}[c]{@{}c@{}}Country\\750km\end{tabular} & \begin{tabular}[c]{@{}c@{}}Continent\\2500km\end{tabular} & \begin{tabular}[c]{@{}c@{}}Street\\1km\end{tabular} & \begin{tabular}[c]{@{}c@{}}City\\25km\end{tabular} & \begin{tabular}[c]{@{}c@{}}Region\\200km\end{tabular} & \begin{tabular}[c]{@{}c@{}}Country\\750km\end{tabular} & \begin{tabular}[c]{@{}c@{}}Continent\\2500km\end{tabular}  \\ 
\midrule
\text{[L]kNN}, sigma=4~\cite{vo2017revisiting} & ICCV'17                    & 7.2                                                 & 19.4                                               & 26.9                                                  & 38.9                                                   & 55.9                                                      & 2.3                                                 & 5.7                                                & 11                                                    & 23.5                                                   & 42                                                         \\
PlaNet~\cite{weyand2016planet}         & ECCV'16                    & 8.5                                                 & 24.8                                               & 34.3                                                  & 48.4                                                   & 64.6                                                      & 5.6                                                 & 14.3                                               & 22.2                                                  & 36.4                                                   & 55.8                                                       \\
CPlaNet~\cite{seo2018cplanet}       & ECCV'18                    & 10.2                                                & 26.5                                               & 34.6                                                  & 48.6                                                   & 64.6                                                      & 7.9                                                 & 14.8                                               & 21.9                                                  & 36.4                                                   & 55.5                                                       \\
ISNs~\cite{muller2018geolocation}           & ECCV'18                    & 10.5                                                & 28                                                 & 36.6                                                  & 49.7                                                   & 66                                                        & 6.5                                                 & 16.2                                               & 23.8                                                  & 37.4                                                   & 55                                                         \\
Translocator~\cite{pramanick2022world}   & ECCV'22                    & 11.8                                                & 31.1                                               & 46.7                                                  & 58.9                                                   & 80.1                                                      & 8.4                                                 & 18.6                                               & 27                                                    & 41.1                                                   & 60.4                                                       \\
GeoDecoder~\cite{clark2023we}     & ICCV'23                    & 12.8                                                & 33.5                                               & 45.9                                                  & 61                                                     & 76.1                                                      & 10.3                                                & 24.4                                               & 33.9                                                  & 50                                                     & 68.7                                                       \\
GeoCLIP~\cite{vivanco2023geoclip}        & NeurIPS'23                 & 14.11                                               & 34.47                                              & 50.65                                                 & 69.67                                                  & 83.82                                                     & 9.59                                                & 19.31                                              & 32.63                                                 & 55                                                     & 74.69                                                      \\
Img2Loc~\cite{zhou2024img2loc}        & SIGIR'24                   & 15.34                                               & 39.83                                              & 53.59                                                 & 69.7                                                   & 82.78                                                     & 19.78                                       & 30.71                                      & 41.4                                          & 58.11                                                  & 74.07                                                      \\
PIGEON~\cite{haas2024pigeon}         & CVPR'24                    & 11.3                                                & 36.7                                               & 53.8                                                  & \uline{72.4}                                           & \uline{85.3}                                              & 10.4                                                & 23.7                                               & 40.6                                                  & 62.2                                           & 77.7                                               \\
G3~\cite{jia2024g3}              & NeurIPS'24                 & \uline{16.65}                                       & \uline{40.94}                                      & \uline{55.56}                                         & 71.24                                                  & 84.68                                                     & \uline{23.99}                                      & \uline{35.89}                                     & \uline{46.98}                                        & \uline{64.26}                                         & \uline{78.15}                                             \\ 
\midrule
\rowcolor{results} \textbf{GeoRanker}            &       Ours                     & \textbf{18.79}                                      & \textbf{45.05}                                     & \textbf{61.49}                                        & \textbf{76.31}                                         & \textbf{89.29}                                            &      \textbf{32.94}                                               &    \textbf{43.54}                                                &    \textbf{54.32}                                                   &    \textbf{69.79}                                                    &      \textbf{82.45}                                                      \\ \rowcolor{results}
Rel. Improvement        &      $\Delta$                      & $\uparrow12.9\%$                                     & $\uparrow10.0\%$                                    & $\uparrow10.7\%$                                       & $\uparrow5.4\%$                                        & $\uparrow4.7\%$                                           &      $\uparrow37.3\%$                                               &      $\uparrow21.3\%$                                              &   $\uparrow15.6\%$                                                    &  $\uparrow8.6\%$                                                      &   $\uparrow5.5\%$                                                         \\
\bottomrule
\end{tabular}}
\end{table}

\subsection{Ablation Study}
\label{sec:ablation}
To better understand the contribution of each component, we conduct ablation studies by systematically varying key modules of our approach. (1) \textbf{w/o $\mathcal{L}_{\text{PL}}^{(2)}$.} Our method without second-order distance loss in training. (2) \textbf{w/o $\mathcal{C}_{\text{neg}}$.} Our method without negative information in training and inference. (3) \textbf{w/o $c_m^{\text{text}}$.} Our method without textual descriptions of candidates in training and inference. (4) \textbf{w/o $c_m^{\text{img}}$.} Our method without image data of candidates in training and inference. (5) \textbf{w/o $\mathcal{C}_g$.} Our method without generated candidates in inference.
\setlength{\intextsep}{0pt}
\setlength{\columnsep}{10pt}
\begin{wraptable}{r}{0.5\textwidth}
\centering
\caption{Ablation study on IM2GPS3K.}
\label{tab:ablation_study}
\resizebox{\linewidth}{!}{
\begin{tabular}{lccccc} 
\toprule
Methods                                    & \begin{tabular}[c]{@{}c@{}}Street\\1km\end{tabular} & \begin{tabular}[c]{@{}c@{}}City\\25km\end{tabular} & \begin{tabular}[c]{@{}c@{}}Region\\200km\end{tabular} & \begin{tabular}[c]{@{}c@{}}Country\\750km\end{tabular} & \begin{tabular}[c]{@{}c@{}}Continent\\2500km\end{tabular}  \\ 
\midrule
$\text{w/o} \mathcal{L}_{\text{PL}}^{(2)}$ & \uline{18.48}                                       & \uline{44.61}                                      & \uline{60.96}                                         & 75.61                                                   & 88.28                                                      \\
$\text{w/o} \mathcal{C}_{\text{neg}}$      & 17.35                                               & 44.51                                              & 60.82                                                 & \uline{76.37}                                                     & 88.28                                                      \\
$\text{w/o} c_m^{\text{text}}$             & 18.02                                               & 43.91                                              & 60.19                                                 & \textbf{76.61}                                          & 88.62                                                      \\
$\text{w/o} c_m^{\text{img}}$              & 15.58                                               & 41.77                                               & 59.15                                                 & 75.40                                                   & 88.35                                                      \\
$\text{w/o} \mathcal{C}_g$                 & 18.21                                               & 43.47                                              & 59.69                                                 & 75.47                                                  & \uline{88.75}                                              \\ 
\midrule
Ours                                       & \textbf{18.79}                                      & \textbf{45.05}                                     & \textbf{61.49}                                        & 76.31                                         & \textbf{89.29}                                             \\
\bottomrule
\end{tabular}}
\end{wraptable}
Table~\ref{tab:ablation_study} presents an ablation study on the IM2GPS3K dataset, and the results for YFCC4K are illustrated in \textbf{Appendix~\ref{sec:appendix_ablation}}. From Table~\ref{tab:ablation_study} we draw several key insights: 
(1) All components in our framework contribute positively to the final performance, demonstrating the effectiveness of our design. 
(2) Comparing our full model with the variant without second-order distance loss ($\mathcal{L}_{\text{PL}}^{(2)}$), we observe more substantial improvements at coarse-grained levels (e.g., country and continent). This highlights the benefit of modeling second-order spatial relationships among candidates, which enables finer-grained ranking and enhances geolocalization accuracy. 
(3) Removing any of the modality-aware prompt components—such as negative candidates ($\mathcal{C}_{\text{neg}}$), textual descriptions ($c_m^{\text{text}}$), or image data ($c_m^{\text{img}}$)—leads to performance drops, confirming that incorporating multi-modal cues into the prompt is beneficial. Among these, visual information yields the most significant gain, underscoring the importance of image semantics. 
(4) Finally, the variant without generated candidates ($\mathcal{C}_g$) underperforms our method, showing that generated candidates provide complementary value. This is especially important in scenarios where the retrieval database lacks relevant examples, and generation can introduce novel, informative candidates that enhance the overall candidate pool.

\subsection{Hyperparameter Analysis}

\begin{figure}[b]
    \centering
    \includegraphics[width=\linewidth]{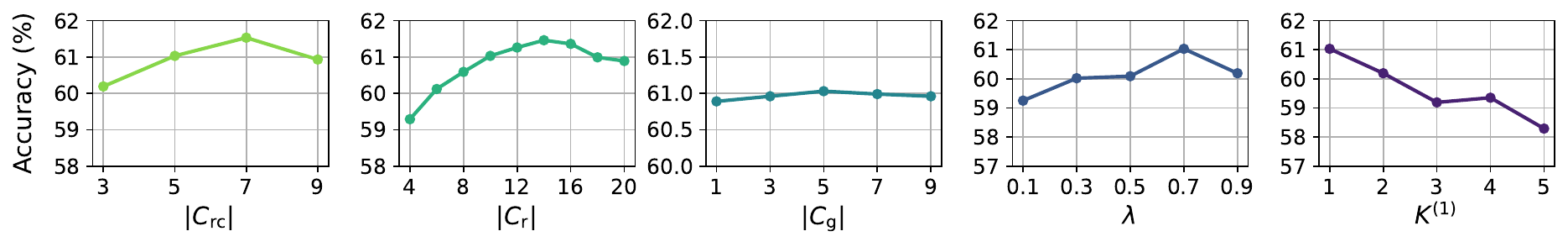}
    \caption{Hyperparameter analysis at the region level on IM2GPS3K. Trends observed at the region level are representative across different geographic levels. Results for all hyperparameters across all levels can be found in \textbf{Appendix~\ref{sec:appendix_hyper_analysis}}. }
    \label{fig:hyper}
\end{figure}

To better understand the impact of key hyperparameters, we conduct a systematic ablation study by varying each at a time while keeping others fixed. The hyperparameters considered include: (1) \textbf{Number of retrieval candidates in training $\rvert \mathcal{C}_{{rc}} \rvert$.} (2) \textbf{Number of retrieval candidates during inference $\rvert \mathcal{C}_{{r}} \rvert$.} (3) \textbf{Number of generated candidates during inference $\rvert \mathcal{C}_{{g}} \rvert$.} (4) \textbf{Weighting coefficient $\lambda$.} (5) \textbf{Number of top elements involved in the first-order loss ($K^{(1)})$.} Unless otherwise specified, we use the following default values: $\rvert \mathcal{C}_{{rc}} \rvert=5$, $\rvert \mathcal{C}_{{r}} \rvert=10$, $\rvert \mathcal{C}_{{g}} \rvert=5$, $\lambda_1=0.7$, and $K^{(1)}=1$.
Empirically, we find that the trends of each hyperparameter remain largely consistent across all levels, indicating the stability and robustness of our model under varying geolocalization granularities. For clarity and brevity, we present results at the region level as a representative example in Figure~\ref{fig:hyper}, while full results for all levels are provided in \textbf{Appendix~\ref{sec:appendix_hyper_analysis}}. In addition, an ablation study on the retrieval pool size is provided in \textbf{Appendix~\ref{sec:retrieval_pool}} for reference.

\textbf{Impact of candidate scales in training and inference:}
(1) Increasing $| \mathcal{C}_{\text{rc}} |$ initially improves performance, followed by a plateau. This suggests that increasing the number of candidates moderately raises task difficulty, which in turn provides more supervision signals and benefits model training.
(2) The number of retrieval candidates used during inference $\rvert \mathcal{C}_{\text{r}} \rvert$ also exhibits a rising-then-stabilizing trend. A larger pool of retrieval candidates increases the likelihood of including the correct candidate, and the consistent performance gain further demonstrates the effectiveness of our GeoRanker, which can robustly identify the most relevant one from a diverse set.
(3) The model shows relatively flat performance when varying $| \mathcal{C}_{\text{g}} |$, indicating that even a small number of generated candidates is sufficient to yield competitive performance.

\textbf{Impact of hyperparameters in multi-order distance objective:}
(1) As the weighting factor $\lambda$ increases, performance first improves and then declines. This highlights a trade-off between the first- and second-order objectives—overweighting the former can reduce the benefit of modeling relative spatial relationships.
(2) $K^{(1)}$ shows a consistent downward trend. Larger values introduce more candidate combinations in the partial PL loss during training, which may deviate from the candidate distribution at inference and lead to train-test mismatch. This weakens the supervision signal and degrades performance.

\subsection{Comparison with Other Ranking Baselines} \label{sec:ranking_baselines}
\begin{figure}
\centering
\begin{minipage}{0.5\linewidth}
\centering
\captionof{table}{Comparison with other ranking baselines.}
\label{tab:ranking_baseline}
\resizebox{\linewidth}{!}{
\begin{tabular}{cccccc} 
\toprule
\multirow{2}{*}{Methods} & \multicolumn{5}{c}{IM2GPS3K}                                                                                                                                                                                                                                                           \\ 
\cmidrule{2-6}
                         & \begin{tabular}[c]{@{}c@{}}Street\\1km\end{tabular} & \begin{tabular}[c]{@{}c@{}}City\\25km\end{tabular} & \begin{tabular}[c]{@{}c@{}}Region\\200km\end{tabular} & \begin{tabular}[c]{@{}c@{}}Country\\750km\end{tabular} & \begin{tabular}[c]{@{}c@{}}Continent\\2500km\end{tabular}  \\ 
\midrule
Random                   & 10.04                                               & 29.72                                              & 42.17                                                 & 57.82                                                  & 75.24                                                      \\
Top1                     & 13.31                                               & 34.03                                              & 45.48                                                 & 61.56                                                  & 78.04                                                      \\
Prompting                & \uline{16.62}                                       & \uline{40.21}                                      & \uline{54.55}                                         & \uline{70.07}                                          & \uline{83.24}                                              \\ 
\midrule
Ours                     & \textbf{18.79}                                      & \textbf{45.05}                                     & \textbf{61.49}                                        & \textbf{76.31}                                         & \textbf{89.29}                                             \\
\bottomrule
\end{tabular}}
\end{minipage}
\hspace{0.04\linewidth}
\begin{minipage}{0.28\linewidth}
\centering
\includegraphics[width=\linewidth]{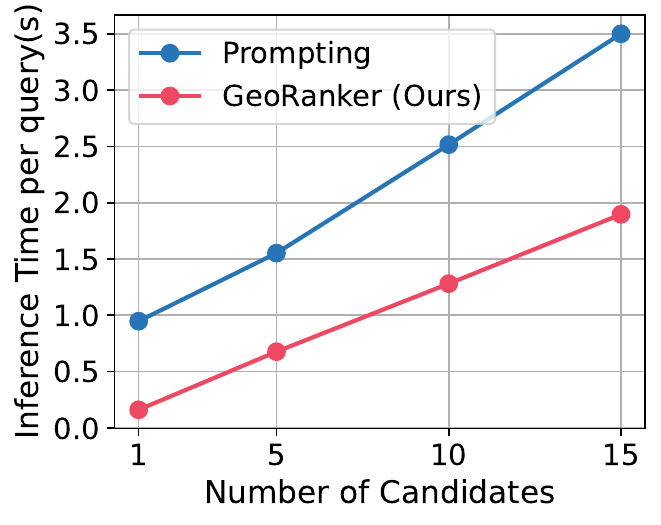}
\captionof{figure}{Time efficiency.}
\label{fig:time_efficiency}
\end{minipage}
\vspace{-1em}
\end{figure}

To demonstrate the superiority of GeoRanker in ranking ability, we conduct comparative experiments with the following ranking baselines. (1)~\textbf{Random:} Randomly sampling one candidate from $\mathcal{C}_{\text{r}} \cup \mathcal{C}_{\text{g}}$ as prediction.
 (2) \textbf{Top-1:} Using embedding similarity to rank candidates and select the top-1 as prediction. The embedding model is fine-tuned following G3~\cite{jia2024g3} with multi-modal information.
    (3) \textbf{Prompting:} The query image, candidate information, and negative samples are incorporated into the prompt, using LVLM to select the most appropriate candidate as the final prediction. We use Qwen2-VL-7b-Instruct for fair comparison. 
From Table~\ref{tab:ranking_baseline}, we can find that our approach (GeoRanker) outperforms all baselines, achieving the highest performance across all metrics. This is because GeoRanker leverages large vision-language models to jointly encode query-candidate interactions and learns fine-grained distance representation through multi-order distance loss during training, enabling it to effectively select accurate predictions from a pool of candidates.

\subsection{Efficiency Analysis}

Beyond accuracy, efficiency is critical for real-world deployment. We evaluate our approach along two dimensions: \textbf{time efficiency}, measuring inference latency, and \textbf{data efficiency}, assessing the effectiveness of data usage.

\textbf{Time efficiency.} 
Figure~\ref{fig:time_efficiency} compares the inference time of GeoRanker with the prompting-based method (introduced in Section~\ref{sec:ranking_baselines}) across varying numbers of candidate inputs. As expected, inference time increases for both methods as the number of candidates grows, due to additional scoring iterations required for GeoRanker and longer prompts for the prompting baseline.
Notably, GeoRanker consistently achieves substantially lower inference latency compared to prompting. Within the 1-10 candidate size range, GeoRanker takes less than half the time required by prompting.
It is also worth highlighting that GeoRanker naturally supports parallel computation over candidate scoring, enabling substantial reductions in inference latency for large-scale deployment. In contrast, Prompting suffers from longer and sequential input construction, which limits such optimization.

\textbf{Data efficiency.}
\begin{figure}
    \centering
    \includegraphics[width=\linewidth]{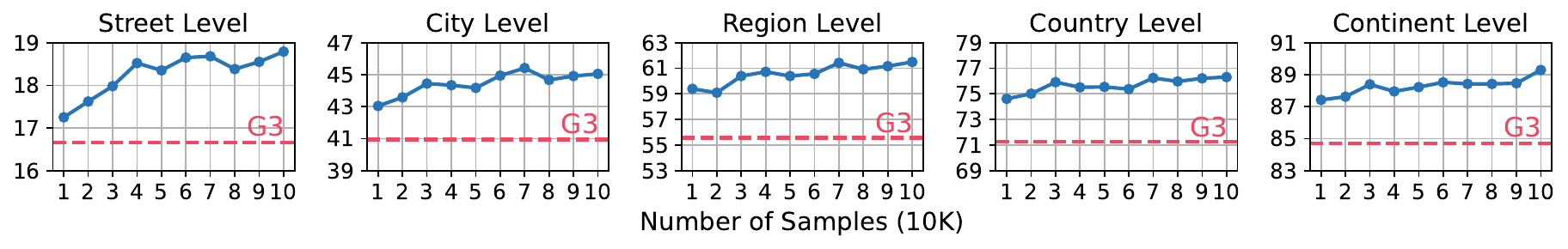}
    \caption{Data efficiency analysis.}
    \label{fig:data_efficiency}
\end{figure}
Figure~\ref{fig:data_efficiency} illustrates the performance of GeoRanker at different geographic scales when fine-tuned with varying amounts of data. The x-axis represents the number of samples (in units of 10K), and the y-axis shows the corresponding accuracy.
From Figure~\ref{fig:data_efficiency}, we observe the following:
(1) GeoRanker exhibits a stable and consistent improvement in accuracy as the training data size increases across all geographic levels, demonstrating strong scalability and generalization capacity.
(2) For comparison, we also plot the performance of the state-of-the-art method G3. Remarkably, GeoRanker surpasses G3 across all levels even when fine-tuned on just 10\% samples, highlighting its data efficiency—the ability to achieve strong performance with limited supervision.

\subsection{Impact of Backbone Scale}

\setlength{\intextsep}{-7pt} 
\setlength{\columnsep}{10pt} 
\begin{wrapfigure}{r}{0.30\textwidth}
\vspace{0mm}
\centering
    \includegraphics[width=\linewidth]{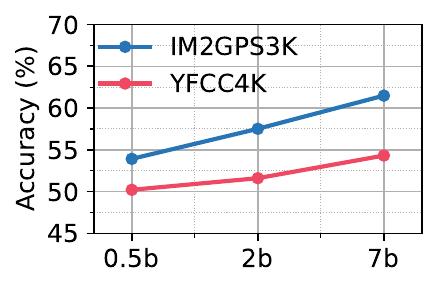}
    \caption{Impact of Backbone Scale on Region Level.}
    \label{fig:scale}
\end{wrapfigure}

To investigate the impact of backbone model scale on performance, we conduct experiments using llava-onevision~\cite{li2024llava} (0.5B) and Qwen2-VL models~\cite{wang2024qwen2} with 2B and 7B parameters.
As shown in Figure~\ref{fig:scale}, and results across all geographic levels in \textbf{Appendix~\ref{sec:appendix_scale}}, GeoRanker's performance consistently improves as the backbone model size increases on both IM2GPS3K and YFCC4K.
These results indicate that GeoRanker benefits from more powerful LVLM backbones and follows the scaling law, suggesting that its upper-bound performance can be further improved with larger models.

\subsection{Case Study}

\begin{figure}
    \centering
    \includegraphics[width=\linewidth]{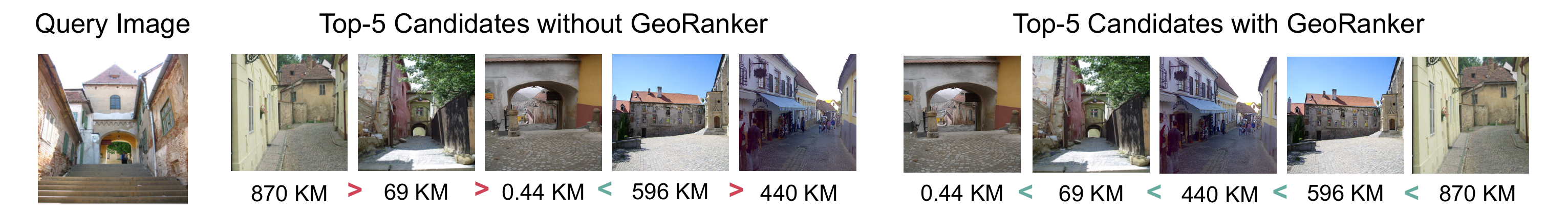}
    \caption{Case study illustrating the effectiveness of GeoRanker in re-ranking candidates.}
    \label{fig:case_study}
\end{figure}

To intuitively demonstrate the effectiveness of GeoRanker, we present a qualitative case study in Figure~\ref{fig:case_study}. As shown on the left, the top-5 candidates retrieved are not well ordered by geographic proximity; visually similar but geographically distant images (e.g., 870 KM away) appear at the top ranks.
After reranking with GeoRanker, the candidates are successfully reordered by their true geographic distances, with the closest image (0.44 km away) ranked at the top and the farthest ones pushed lower in the list.
This result highlights GeoRanker’s ability to model complex spatial relationships through query–candidate interactions, further improving the geolocalization accuracy. 

\section{Conclusion}

In this paper, we propose GeoRanker, a distance-aware ranking framework built upon LVLM. To enhance training, we introduce a novel multi-order distance loss that captures both absolute distances and relative spatial relationships among candidate locations.
To support this framework, we construct GeoRanking, the first dataset specifically designed for spatial ranking tasks.
Extensive experiments on IM2GPS3K and YFCC4K demonstrate the effectiveness of GeoRanker over baselines.

\clearpage
\section*{Acknowledgements}
The authors would like to thank Leitian Tao for the valuable feedback on the work. Pengyue Jia and Xiangyu Zhao are supported by the Institute of Digital Medicine of City University of Hong Kong (No.9229503), Research Impact Fund (No.R1015-23), Collaborative Research Fund (No.C1043-24GF), and General Research Fund (No.11218325).

\bibliography{references}{}
\bibliographystyle{unsrt}

\clearpage
\appendix
\begin{center}
    \LARGE \textbf{Technical Appendix}
    \vspace{1em}
\end{center}
\addcontentsline{toc}{part}{Appendix}

\begingroup
  \etocsetnexttocdepth{subsection}
  \etocsettocstyle{\section*{Table of Contents}}{}
  \localtableofcontents
\endgroup
\clearpage

\section{Prompts}

\textbf{Prompting for generating candidates $\mathcal{C}_g$.} \label{sec:appendix_generation_prompt}
Following previous work~\cite{jia2024g3}, we use the following prompt template for generating candidates: 

\begin{tcolorbox}[colback=gray!5!white,colframe=gray!75!black]
 \textcolor{myblue}{\{query image\}} Suppose you are an expert in geolocalization. You have the ability to give two number GPS coordinates given an image. Please give me the location of the given image. Your answer should be in the following JSON format without any other information: \{"latitude": float,"longitude": float\}.
\end{tcolorbox}

\section{GeoRanking Data Entries} \label{sec:appendix_georanking}

\begin{figure}
    \centering
    \includegraphics[width=\linewidth]{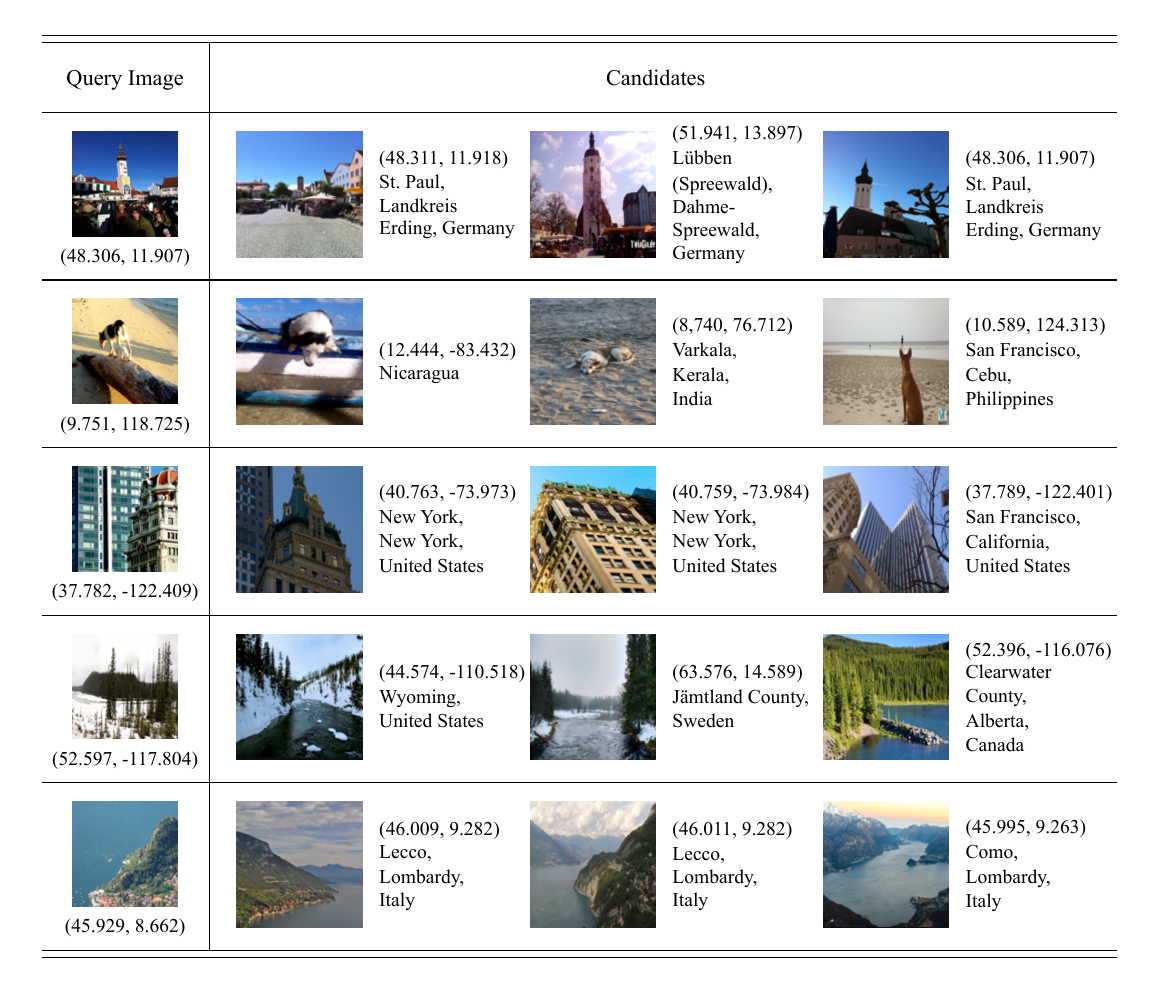}
    \caption{Examples of GeoRanking Data Entries.}
    \label{fig:appendix_georanking_entries}
\end{figure}

Figure~\ref{fig:appendix_georanking_entries} illustrates example entries from the GeoRanking dataset. Specifically, each query image is associated with 20 candidates, and each candidate contains GPS coordinates, textual descriptions, and image data.
In total, GeoRanking includes 100K samples and 2 million query–candidate pairs.
To the best of our knowledge, GeoRanking is the first dataset specifically designed for modeling distance-aware ranking between geographic entities.
We release the dataset publicly and hope it will foster future research in areas such as GeoAI, information retrieval, and vision-language modeling.

\section{More Information on Training and Inference}
\label{sec:appendix_hyper}

\begin{table}
\centering
\caption{More Details on Training and Inference.}
\label{tab:more_hyper}
\resizebox{0.8\linewidth}{!}{
\begin{tabular}{cc} 
\toprule
Parameter            & Setting                                                \\ 
\midrule
GPU                  & NVIDIA L40S * 4                                  \\
Training Time        & 16 hours / epoch                        \\
Total params         &  8,298,256,896                                      \\
Trainable params     &  6,881,280 (0.083\%)                            \\
Dataset Samples      & 100K                                         \\
Batch Size           & 4                                               \\
Batch Size per Device & 1 \\
Training GPU Memory Consumption           &   30 GB / GPU                                             \\
VLM Backbone       & Huggingface Qwen2-VL-7b-Instruct           \\
Deepspeed & Stage 2 \\
\bottomrule
\end{tabular}
}
\end{table}

In this section, we provide additional details regarding the training and inference setup. Table~\ref{tab:more_hyper} summarizes the key hyperparameters used during these phases. Most experiments were conducted on four NVIDIA L40S GPUs. We also performed tests on two NVIDIA H200 GPUs, where training took approximately 7.5 hours per epoch with a batch size of 4, consuming around 90 GB of GPU memory per device with the gradient checkpointing off.

\section{Baseline Method Details}
\label{sec:appendix_baseline}

In this section, we will give introductions to the baselines:
\begin{itemize}[leftmargin=*]
    \item \textbf{$\text{[L]kNN},\sigma=4$}~\cite{vo2017revisiting}. kNN first retrieves the top-$k$ nearest neighbor images and aggregates their coordinates to form the final prediction. As the $k$ decreases, the aggregation process becomes more focused. When $k$ equals 1, the method turns to the NN.
    \item \textbf{PlaNet}~\cite{weyand2016planet}. PlaNet is the first work to formulate the worldwide geolocalization task as a classification problem. It partitions the Earth's surface into a large number of geographical cells and trains a convolutional neural network to predict the correct cell for each image. Unlike previous approaches that primarily rely on landmark recognition or approximate matching with global image descriptors, PlaNet effectively integrates multiple visible cues within the image to enhance localization accuracy.
    \item \textbf{CPlaNet}~\cite{seo2018cplanet}. CPlaNet follows PlaNet and proposes combinatorial partitioning, which generates fine-grained output classes by intersecting larger partitions.
    \item \textbf{ISNs}~\cite{muller2018geolocation}. ISNs enhance the input image information by extracting additional scene context features, such as indoor, natural, or urban environments, alongside the original image content. By incorporating these richer contextual cues, ISNs achieves improved localization performance.
    \item \textbf{Translocator}~\cite{pramanick2022world}. Translocator designs a dual-branch transformer framework that simultaneously ingests the original image and its semantic segmentation map. This architecture enables the extraction of fine-grained spatial cues and the construction of more robust feature representations for geolocalization.
    \item \textbf{GeoDecoder}~\cite{clark2023we}. GeoDecoder identifies that earlier methods insufficiently leverage hierarchical spatial information. It addresses this by proposing a cross-attention mechanism that explicitly captures relationships across heterogeneous features, enhancing the model's ability to interpret complex location-dependent features.
    \item \textbf{GeoCLIP}~\cite{vivanco2023geoclip}. GeoCLIP extends the CLIP architecture by introducing a GPS encoder, aligning geographic coordinates with image and GPS embeddings. This enhancement enables more effective modeling of worldwide geolocalization tasks by incorporating spatial information directly into the learned feature space.
    \item \textbf{Img2Loc}~\cite{zhou2024img2loc}. Img2Loc advances geolocalization by integrating an RAG pipeline. It first retrieves visually similar candidates, then formulates a prompt incorporating these candidates' coordinates, guiding a vision-language model to generate a final prediction.
    \item \textbf{PIGEON}~\cite{haas2024pigeon}. PIGEON introduces an innovative framework that combines semantic geocell partitioning, multi-task contrastive pretraining, and a novel loss function. By clustering candidate locations semantically and refining predictions through targeted retrieval, PIGEON significantly boosts localization accuracy.
    \item \textbf{G3}~\cite{jia2024g3}. G3 proposes a three-stage framework comprising Geo-alignment, Geo-diversification, and Geo-verification. Geo-alignment aligns GPS coordinates, textual descriptions, and visual data into a unified multi-modal representation to strengthen retrieval capabilities. Subsequently, Geo-diversification and Geo-verification are integrated within an RAG framework to robustly generate and select candidate geolocations.
\end{itemize}

\section{Query Images with Different Error Thresholds} \label{sec:appendix_error_thresholds}

\begin{figure}
    \centering
    \includegraphics[width=\linewidth]{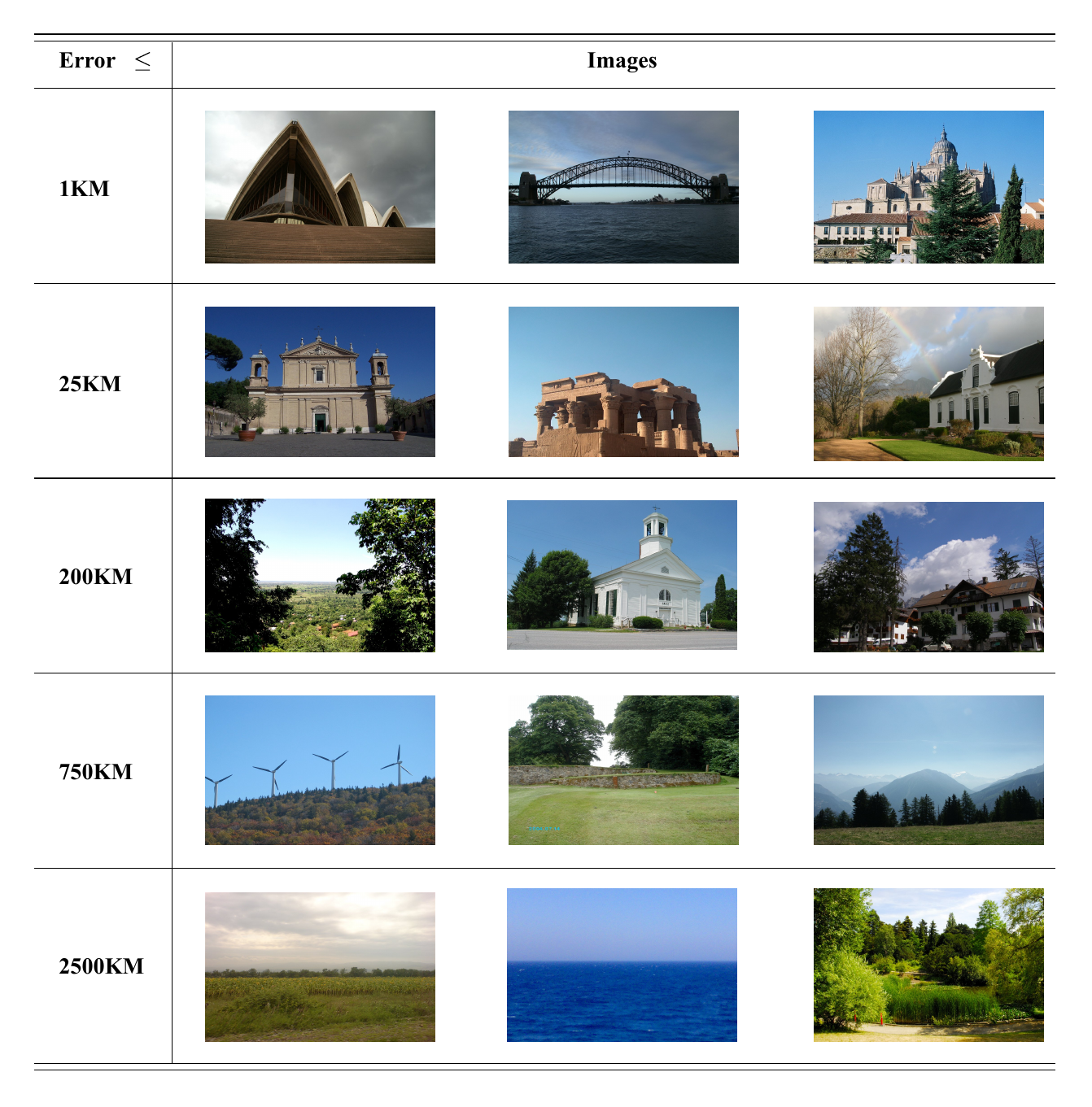}
    \caption{Example query images fall in different error thresholds.}
    \label{fig:appendix_error_thresholds}
\end{figure}

Figure~\ref{fig:appendix_error_thresholds} presents example query images under different error thresholds (1km, 25km, 200km, 750km, and 2500km).
We observe that images with errors within 1km often contain distinctive location cues, such as landmark buildings, which facilitate accurate geolocalization. This is partly because retrieval candidates are more likely to retrieve visually similar images from the database due to the popularity of such locations.
Additionally, generated candidates tend to produce more reliable predictions in these cases, as the locations are well-represented in the world knowledge embedded in large vision-language models.
In contrast, query images with large geolocalization errors (e.g., 2500km) typically lack informative visual cues—such as images depicting open oceans or vast grasslands—making it extremely challenging to infer their true locations. In such cases, neither retrieval nor generation is likely to yield useful candidates.

\section{Experimental Analysis with Ranking Metrics} \label{sec:ranking_metrics}

\begin{table}
\centering
\caption{Ranking performance comparison on IM2GPS3K.}
\label{tab:ranking_metrics}
\resizebox{0.9\textwidth}{!}{
\begin{tabular}{lcccccc} 
\toprule
Methods      & Recall@1        & Recall@5        & Recall@10       & NDCG@5          & NDCG@10         & NDCG@20          \\ 
\midrule
G3 Retrieval & 0.0894          & 0.3217          & 0.5672          & 0.6238          & 0.6735          & 0.7989           \\
GeoRanker    & \textbf{0.1982} & \textbf{0.5169} & \textbf{0.7387} & \textbf{0.8026} & \textbf{0.8419} & \textbf{0.8872}  \\
\bottomrule
\end{tabular}}
\end{table}

To comprehensively assess the ranking capability of GeoRanker as a ranking model, we additionally adopt standard learning-to-rank metrics, specifically Recall@K and NDCG@K. We take the top-20 candidates retrieved by G3 as the candidate pool. For Recall@K, we designate the most accurate candidate (i.e., the one closest to the ground truth) as the ground truth candidate and compute whether the ground truth candidate appears within the top-k predictions. For NDCG@K, we assign relevance scores based on distance to the ground truth: candidates within 1 km are assigned a label of 1.0; those within 1–25 km receive 0.8; 25–200 km receive 0.6; 200–750 km receive 0.4; 750–2500 km receive 0.2; and those beyond 2500 km are labeled 0. The results are shown in Table~\ref{tab:ranking_metrics}: GeoRanker significantly improves both Recall and NDCG over G3. This demonstrates that our ranking module effectively re-orders the candidates to improve fine-grained geolocalization accuracy.

\section{Complete experimental results on ablation study} \label{sec:appendix_ablation}

\begin{table}
\centering
\caption{Complete ablation study on IM2GPS3K and YFCC4K.}
\label{tab:appendix_ablation}
\resizebox{\linewidth}{!}{
\begin{tabular}{lcccccccccc} 
\toprule
\multirow{2}{*}{\textbf{Methods}}                   & \multicolumn{5}{c}{IM2GPS3K}                                                                                                                                                                                                                                                          & \multicolumn{5}{c}{YFCC4K}                                                                                                                                                                                                                                                             \\ 
\cmidrule{2-11}
                                           & \begin{tabular}[c]{@{}c@{}}Street\\1km\end{tabular} & \begin{tabular}[c]{@{}c@{}}City\\25km\end{tabular} & \begin{tabular}[c]{@{}c@{}}Region\\200km\end{tabular} & \begin{tabular}[c]{@{}c@{}}Country\\750km\end{tabular} & \begin{tabular}[c]{@{}c@{}}Continent\\2500km\end{tabular} & \begin{tabular}[c]{@{}c@{}}Street\\1km\end{tabular} & \begin{tabular}[c]{@{}c@{}}City\\25km\end{tabular} & \begin{tabular}[c]{@{}c@{}}Region\\200km\end{tabular} & \begin{tabular}[c]{@{}c@{}}Country\\750km\end{tabular} & \begin{tabular}[c]{@{}c@{}}Continent\\2500km\end{tabular}  \\ 
\midrule
$\text{w/o} \mathcal{L}_{\text{PL}}^{(2)}$ & \uline{18.48}                                       & \uline{44.61}                                      & \uline{60.96}                                         & 75.61                                                  & 88.28                                                     & 31.97                                       & \uline{43.12}                                      & 53.53                                      & 69.03                                          & 81.19                                                      \\
$\text{w/o} \mathcal{C}_{\text{neg}}$      & 17.35                                               & 44.51                                              & 60.82                                                 & \uline{76.37}                                                  & 88.28                                                     & 31.57                                               & 43.06                                              & 53.62                                                 & 69.09                                                  & 81.67                                                      \\
$\text{w/o} c_m^{\text{text}}$             & 18.02                                               & 43.91                                              & 60.19                                                & \textbf{76.61}                                          & 88.62                                                     & 31.70                                               & 43.06                                              & \uline{54.03}                                                 & 69.42                                                  & 82.07                                                      \\
$\text{w/o} c_m^{\text{img}}$              & 15.58                                               & 41.77                                              & 59.15                                                 & 75.40                                                  & 88.35                                                     & 15.81                                               & 27.86                                              & 41.31                                                 & 61.39                                                  & 77.66                                                      \\
$\text{w/o} \mathcal{C}_g$                 & 18.21                                               & 43.47                                              & 59.69                                                 & 75.47                                                  & \uline{88.75}                                             & \uline{32.60}                                               & 43.03                                              & 53.43                                                 & \uline{69.77}                                                  & \textbf{82.71}                                             \\ 
\midrule
Ours                                       & \textbf{18.79}                                      & \textbf{45.05}                                     & \textbf{61.49}                                        & 76.31                                         & \textbf{89.29}                                            & \textbf{32.94}                                      & \textbf{43.54}                                     & \textbf{54.32}                                        & \textbf{69.79}                                         & \uline{82.45}                                             \\
\bottomrule
\end{tabular}}
\end{table}

Table~\ref{tab:appendix_ablation} presents the complete ablation results on both IM2GPS3K and YFCC4K. Consistent with the findings discussed in the main text, we observe that each component in our framework contributes positively to overall performance.
Moreover, different types of contextual information incorporated into the prompt—such as visual cues, textual descriptions, and negative examples—all help improve both model training and inference.
Finally, generated candidates are shown to complement retrieval-based candidates effectively. This is particularly beneficial for rare or long-tail query images, where retrieval candidates alone may fail to provide sufficient clues for accurate geolocation.

\section{Hyperparameter Analysis with All Geographic Levels} \label{sec:appendix_hyper_analysis}

\begin{figure}
    \centering
    \includegraphics[width=\linewidth]{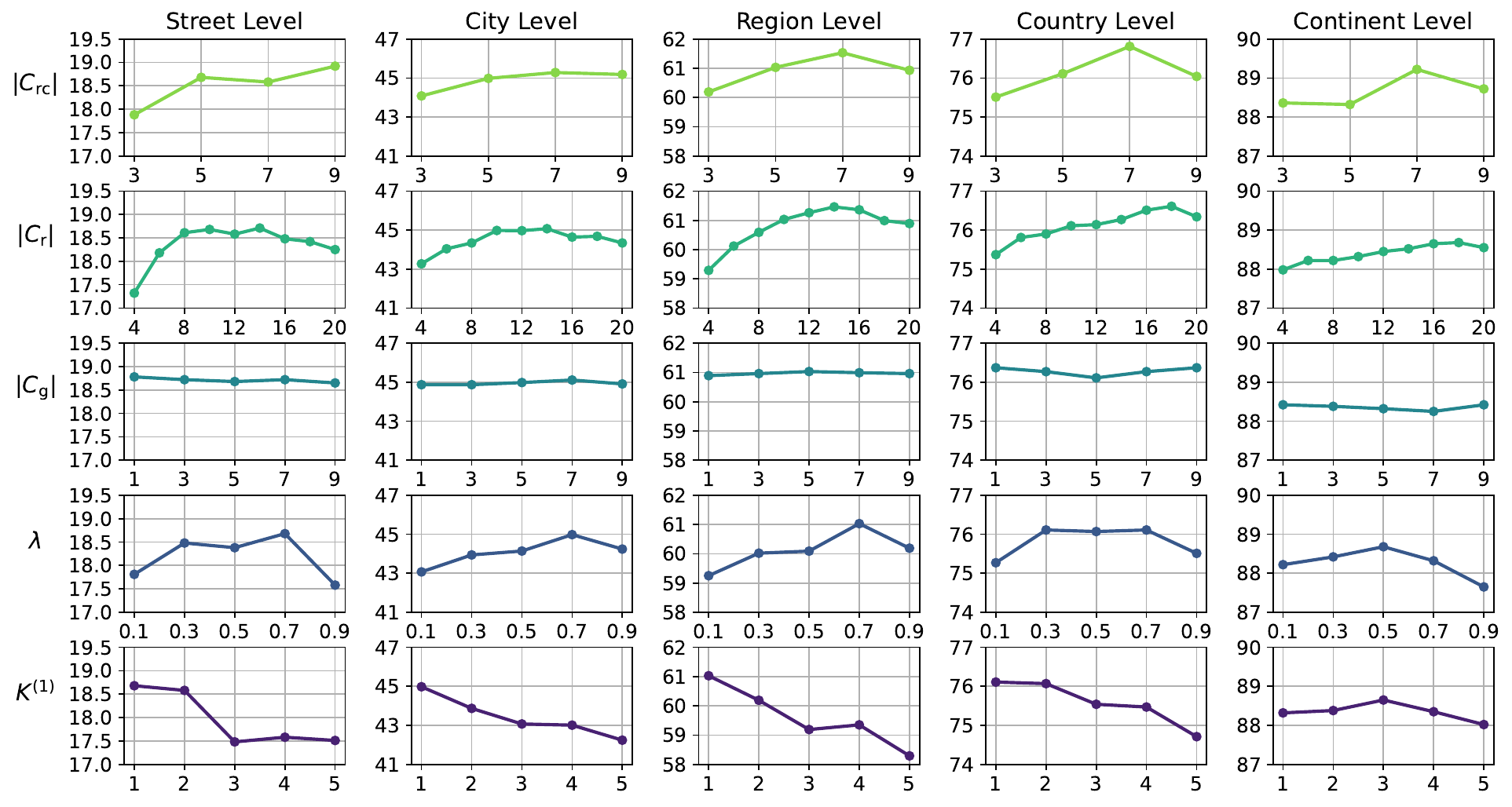}
    \caption{Hyperparameter analysis with all geographic levels on IM2GPS3K.}
    \label{fig:hyper-appendix}
\end{figure}

Figure~\ref{fig:hyper-appendix} shows the impact of different hyperparameters on GeoRanker across all geographic levels. As observed, the trends of each hyperparameter remain largely consistent across levels, highlighting the stability and robustness of our model under varying localization granularities.

\section{Ablation Study on Retrieval Pool Size} \label{sec:retrieval_pool}

\begin{table}
\centering
\caption{Effect of retrieval pool size on geolocalization accuracy.}
\label{tab:retrieval_pool}
\resizebox{0.6\textwidth}{!}{
\begin{tabular}{lccccc} 
\toprule
Sample Ratio & 1km   & 25km  & 200km & 750km & 2500km  \\ 
\midrule
10\%         & 14.91 & 40.77 & 58.26 & 75.61 & 88.52   \\
25\%         & 16.95 & 43.04 & 59.93 & 76.38 & 88.59   \\
50\%         & 17.25 & 43.31 & 60.06 & 76.74 & 88.72   \\
100\%        & 18.79 & 45.05 & 61.49 & 76.31 & 89.29   \\
\bottomrule
\end{tabular}}
\end{table}

To investigate the effect of retrieval pool size on the performance of GeoRanker during inference, we conduct experiments on the IM2GPS3K dataset. Specifically, we sample 10\%, 25\%, and 50\% of the full retrieval pool and compare their prediction accuracy. From Table~\ref{tab:retrieval_pool}, we can draw the following conclusions: (1) GeoRanker’s performance consistently improves as the retrieval pool size increases; (2) The improvement is more pronounced on fine-grained metrics (1km, 25km, 200km) compared to coarse-grained ones (750km, 2500km), indicating that a larger pool provides more precise candidates that benefit high-resolution geolocalization.

\section{Complete Experimental Results on Backbone Model Scale} \label{sec:appendix_scale}

\begin{figure}
    \centering
    \includegraphics[width=\linewidth]{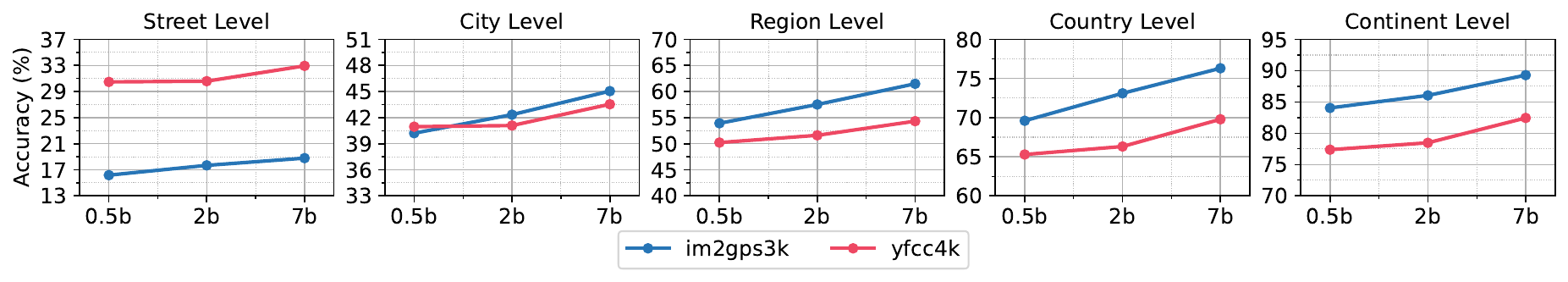}
    \caption{Impact of Backbone Scale across All Levels.}
    \label{fig:appendix_scale}
\end{figure}

Figure~\ref{fig:appendix_scale} shows the effect of backbone model size across all geographic levels. Consistent performance improvements are observed on both IM2GPS3K and YFCC4K datasets as the backbone scales from 0.5B to 7B parameters, further confirming GeoRanker's scalability and compatibility with stronger LVLM.

\section{Limitations} \label{sec:appendix_limitation}
Our method achieves notable improvements in geolocalization accuracy over existing baselines. In addition, it demonstrates superior time efficiency compared to LVLM prompting methods, and its data efficiency allows strong performance even with relatively limited supervision. However, compared to direct embedding-based retrieval approaches, GeoRanker introduces an additional ranking stage, which leads to increased computational overhead during inference. One solution is to analyze the retrieval results: if the top-k candidates are already geographically concentrated, the ranking step can be skipped without significant loss in accuracy, thereby reducing the overall inference time. In addition, GeoRanker supports parallel scoring of candidates during large-scale deployment, which can significantly improve runtime and computational efficiency.

\end{document}